\documentclass[sigconf,authorversion,nonacm]{acmart}

\AtBeginDocument{%
  \providecommand\BibTeX{{%
    \normalfont B\kern-0.5em{\scshape i\kern-0.25em b}\kern-0.8em\TeX}}}

\setcopyright{acmcopyright}
\copyrightyear{2024}
\acmYear{2024}
\acmDOI{10.1145/xxxxxxx.xxxxxxx}

\acmConference[WiSec '24]{WiSec '24: 17th ACM Conference on Security and Privacy in Wireless and Mobile Networks}{May 27--May 30, 2024}{Seoul, Korea}
\acmBooktitle{WiSec '24: 17th ACM Conference on Security and Privacy in Wireless and Mobile Networks, May 27--May 30, 2024, Seoul, Korea}
\acmPrice{15.00}
\acmISBN{978-1-4503-XXXX-X/21/06}

\usepackage{url}
\usepackage{algorithmic}
\usepackage{graphicx}
\usepackage{textcomp}
\usepackage{xcolor}
\usepackage{comment}
\usepackage{subfigure}
\newtheorem{definition}{Definition}
\usepackage{siunitx}

\usepackage{array}
\newcolumntype{P}[1]{>{\centering\arraybackslash}p{#1}}
\newcommand{\jhc}{\textcolor{red}} 
\newcommand{\dian}{\textcolor{blue}}

\usepackage[utf8]{inputenc}
\usepackage{enumitem}
\setlist[itemize]{leftmargin=*, noitemsep}
\setlist[enumerate]{leftmargin=*, noitemsep}

\begin{document}

\title{SusFL: Energy-Aware Federated Learning-based Monitoring for Sustainable Smart Farms}

\author{Dian Chen}
\email{dianc@vt.edu}
\orcid{0009-0000-7641-454X}
\affiliation{%
  \institution{Virginia Tech}
  \streetaddress{7054 Haycock Rd}
  \city{Falls Church}
  \state{Virginia}
  \country{USA}
  \postcode{22043}
}

\author{Paul Yang}
\affiliation{%
  \institution{Virginia Tech}
  \city{Blacksburg}
  \state{Virginia}
  \country{USA}}
\email{yangaapaul@vt.edu}

\author{Ing-Ray Chen}
\affiliation{%
  \institution{Virginia Tech}
  \streetaddress{7054 Haycock Rd}
  \city{Falls Church}
  \state{Virginia}
  \country{USA}}
\email{irchen@vt.edu}

\author{Dong Sam Ha}
\affiliation{%
  \institution{Virginia Tech}
  \city{Blacksburg}
  \state{Virginia}
  \country{USA}}
\email{dha@vt.edu}

\author{Jin-Hee Cho}
\affiliation{%
  \institution{Virginia Tech}
  \streetaddress{7054 Haycock Rd}
  \city{Falls Church}
  \state{Virginia}
  \country{USA}}
\email{jicho@vt.edu}

\begin{abstract}
We propose a novel energy-aware federated learning (FL)-based system, namely {\tt SusFL}, for sustainable smart farming
to address the challenge of inconsistent health monitoring due to fluctuating energy levels of solar sensors. 
This system equips animals, such as cattle, with solar sensors with computational capabilities, including Raspberry Pis, to train a local deep-learning model on health data. These sensors periodically update Long Range (LoRa) gateways, forming a wireless sensor network (WSN) to detect diseases like mastitis.  Our proposed {\tt SusFL} system incorporates {\em mechanism design}, a game theory concept, for intelligent client selection to optimize monitoring quality while minimizing energy use. This strategy ensures the system's sustainability and resilience against adversarial attacks, including data poisoning and privacy threats, that could disrupt FL operations.  Through extensive comparative analysis using real-time datasets, we demonstrate that our FL-based monitoring system significantly outperforms existing methods in prediction accuracy, operational efficiency, system reliability (i.e., mean time between failures or MTBF), and social welfare maximization by the mechanism designer. Our findings validate the superiority of our system for effective and sustainable animal health monitoring in smart farms. The experimental results show that {\tt SusFL} significantly improves system performance, including a $10\%$ reduction in energy consumption, a $15\%$ increase in social welfare, and a $34\%$ rise in Mean Time Between Failures (MTBF), alongside a marginal increase in the global model's prediction accuracy. 
\end{abstract}



\maketitle

\section{Introduction}

\subsection{Motivation \& Research Goal} \label{subsec:motivation-goal}

Smart farm technologies have shown a revolutionary shift in agricultural practices to meet the increasing global food demand, enhancing resource efficiency and promoting sustainability. This shift is propelled by integrating cutting-edge technologies, such as the Internet-of-Things (IoT), wireless sensors, artificial intelligence (AI), and data analytics. These technologies facilitate real-time monitoring, enable data-driven decision-making, and automate processes to boost productivity, conserve resources, and improve operational efficiency.  In the realm of smart farming, each animal is equipped with a solar-powered sensor that gathers data on the animal's condition. This data is periodically sent to edge devices, such as gateways, via Long-Range (LoRa) transmission and subsequently to a cloud server. This infrastructure allows farmers to efficiently monitor their livestock and manage farm operations. Although the adoption of solar-powered sensors presents labor-saving and environmental benefits, it also introduces significant challenges. Chief among these is the need to maintain a delicate balance between system performance and the sustainability of the wireless sensors' energy levels. Addressing this balance is crucial for the successful implementation and long-term viability of smart farming technologies.

To advance the automation capabilities of smart farming systems, a cloud server can apply a deep learning (DL) model to analyze data transmitted from gateways to detect diseases like mastitis in cattle. This approach acknowledges the potential of federated learning (FL) to enhance disease prediction in smart farms while preserving data privacy and supporting sustainable agricultural practices. FL's efficacy, particularly in terms of prediction accuracy, is largely influenced by the active participation of local devices, such as sensors.  In smart farming environments, the limited energy supply of sensors -- restricted by the size of their solar panels -- poses a significant challenge. This limitation affects not only the local training capabilities of these devices but also the overall performance of the FL system. In addition, the integration of sophisticated technologies increases the susceptibility of smart farms to cyber threats, including attacks that aim at compromising data integrity and disrupting the learning process in AI models.  To counter these issues, we introduce a robust federated learning algorithm, {\tt SusFL}, specifically designed for smart farms to optimize the prediction accuracy of animal diseases and prolong the lifespan of the sensor network, while providing resilience against cyber threats. This innovative approach represents a strategic advancement in maintaining the balance between operational efficiency, system sustainability, and security in smart farming systems.

\subsection{Key Contributions} \label{subsec:contributions}

Our work made the following {\bf key contributions}: 
\begin{enumerate}
\item {\bf Sustainable FL with energy-efficient client selection via mechanism design}: We adopt a strategy rooted in game theory, specifically mechanism design, to enhance the sustainability of smart farming systems. This approach focuses on selecting clients (i.e., sensor-equipped animals) in an energy-efficient manner to optimize the system's overall sustainability. The effectiveness of this method is quantitatively assessed using a reliability metric known as Mean-Time-Between-Failures (MTBF) to build a smart farm that not only conserves energy but also maintains high operational reliability.

\item {\bf Pioneering FL for disease detection in smart farm animals}: Our work represents the inaugural exploration into FL-based monitoring systems designed specifically for smart farms by identifying illnesses in livestock. Unlike previous studies, which did not apply FL for animal disease detection, our approach leverages comprehensive experiments utilizing data from the Internet of Animal Health Things (IoAHT). These experiments are focused on clinical mastitis in cows, thereby providing a solid foundation for validating the effectiveness of FL in real-world agricultural settings.

\item {\bf Robust hierarchical FL under adversarial attacks}: Our research addresses adversarial attacks in smart farm systems using a hierarchical FL framework to maintain high global prediction accuracy for animal health through data quality-aware aggregation. Unlike previous studies, we evaluate the effect of attacks on both prediction accuracy and sensor node energy efficiency in a resource-constrained environment. Utilizing real-world data from Virginia Tech's smart farm, we simulate authentic conditions to validate our approach.
\item {\bf Experimental Validation of {\tt SusFL}}: Our results demonstrate {\tt SusFL}'s enhanced performance, achieving a 10\% decrease in energy use, 15\% boost in social welfare, 34\% higher MTBF, and slightly improved prediction accuracy in the global model.
\end{enumerate}

\section{Background \& Related Work} \label{sec:background-rw}

In this section, we introduce the fundamentals of federated learning (FL) and discuss existing approaches of FL-based energy-efficient systems and monitoring solutions in IoT-based smart environments such as smart farm systems. 

\subsection{Federated Learning}
\begin{figure}
    \centering  \includegraphics[width=0.48\textwidth]{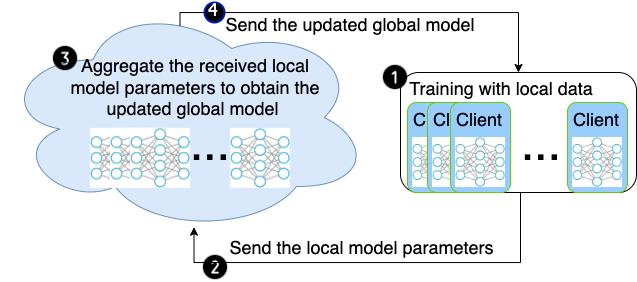}
    \caption{Key steps in FL training process} 
    \label{fig:fl-process}
\end{figure}
FL emerges as a cutting-edge machine learning (ML) paradigm that facilitates collaborative model training across multiple data providers, aiming to construct a high-quality, centralized model without compromising data privacy~\cite{jakub16}. As illustrated in Figure~\ref{fig:fl-process}, the FL framework encompasses a central server hosting the global model and numerous client devices, each maintaining a local model. Within this ecosystem, we consider $N$ distinct data providers, denoted as $\{C_1,\ldots, C_N\}$, each possessing a unique dataset $\{D_1, \ldots, D_N\}$. The training of an ML model $M_{FED}$ under the FL protocol involves the collective effort of all participating data providers. Here, each provider $C_i$ exclusively accesses its dataset $D_i$ to contribute to the global learning process~\cite{yang19}.

The performance of an FL algorithm is quantified in terms of $\delta$-accuracy loss, formally expressed as:
\begin{equation}
|P_{FED} - P_{SUM}| < \delta, \quad \delta \in \mathbb{R},
\end{equation}
where $P_{FED}$ represents the performance of the model trained via the FL algorithm, and $P_{SUM}$ signifies the performance achievable if the model were trained on the aggregated dataset, $D = D_1 \cup \ldots \cup D_N$. This measure of accuracy loss $\delta$, a real number, serves as a critical benchmark to assess the efficiency and effectiveness of the FL approach in approximating the ideal outcome achieved through conventional centralized training methods.

FL systems are categorized based on the distribution characteristics of the data, primarily into two types: horizontal FL (HFL) and vertical (VFL)~\cite{nguyen2021federated}. HFL applies to scenarios where data providers possess datasets that overlap in feature space but are distinct in sample space. Conversely, VFL is appropriate for situations where the datasets share a common sample space but diverge in feature space. Additionally, federated transfer learning (FTL) addresses the more complex scenarios where data holders' datasets vary both in feature and sample spaces.

To advance data privacy and mitigate communication latency within FL methodologies, the concept of {\em hierarchical FL} (HierFL) has emerged as a significant innovation~\cite{liu20}.  Traditional FL architectures are characterized by a two-layer structure, often facing scalability challenges. In response, \citet{liu20} proposed augmenting the FL framework with an additional layer, thereby introducing an intermediary phase of {\em edge server aggregation} between clients and cloud aggregation. This phase acts as a conduit for data processing between the initial client contributions and the final cloud aggregation. By integrating hierarchical FL, we aim to construct a monitoring system for sustainable smart farms that leverages this three-tiered architecture for improved efficiency, enhanced data privacy, and reduced communication delays.

\subsection{Energy-Efficient FL for Smart IoT Systems}

\citet{yang2020energy} developed an energy-efficient Federated Learning (FL) algorithm for latency-sensitive environments like WSNs, optimizing factors like time allocation and computation frequency to minimize energy consumption and FL completion time. Similarly, \citet{pham2022energy} proposed an Energy-efficient FL (E2FL) algorithm tailored for UAVs, addressing nonconvex optimization challenges by breaking them into manageable subproblems.  Additionally, to address the challenge of non-identical independent distribution in efficient FL, \citet{kim2021autofl} introduced AutoFL, utilizing reinforcement learning (RL) to select optimal devices for FL participation, thus enhancing the training process efficiency. \citet{do2021deep} utilized a deep RL (DRL)-based framework to optimize long-term FL performance in environments constrained by resources. They deployed DRL agents on each UAV to dynamically select the best combinations of location, computing speed, transmit power, and bandwidth to maintain energy usage within set budgets for each user.  Moreover, \citet{wang2020federated} innovated an FL scheme using a support vector machine (SVM) to determine the best user association, power allocation, and task allocation strategies to minimize energy usage. \citet{li2021talk} focused on optimizing energy consumption in mobile devices constrained by battery life. This approach balanced local training and data transmission effectively, ensuring the FL model's convergence while preserving the system's overall energy integrity.

Despite the existing solutions for energy-efficient FL in resource-constrained environments above, current methodologies exhibit a significant gap as follows. First of all, they do not account for dynamic and fluctuating energy conditions, such as those encountered with solar-powered sensors. Further, there is a lack of strategies that optimize the FL process through client selection based on a valuation framework that includes criteria like energy level and data quality. This oversight presents a critical opportunity for innovation, suggesting the need for a more adaptive and nuanced approach to FL in environments where energy availability varies significantly over time and across devices. By integrating client selection mechanisms that consider the dynamic energy states and the inherent data quality of participating devices, our work aims to significantly enhance the efficiency and effectiveness of FL systems, especially in scenarios reliant on renewable energy sources.

\subsection{FL-based Monitoring Systems}

\citet{ngu19} introduced a security monitoring system tailored for IoT devices, utilizing an FL framework to identify anomalies within IoT networks based on device-specific communication patterns. Similarly, \citet{sun20} implemented segmented FL within a Local Area Network (LAN) Security monitoring system, enabling local models to adapt to network dynamics. While both studies demonstrated their approaches through extensive experimentation, they notably lacked a comprehensive set of metrics to adequately assess the performance of their monitoring systems.

\citet{wu22} developed {\em FedHome}, an FL-based health monitoring system designed to predict human conditions accurately while ensuring data privacy. They used a generative convolutional autoencoder (GCAE) to minimize the computational and communication overhead of parameter exchange between cloud servers and edge devices. In a similar vein, \citet{khoa21} employed an autoencoder (AE) that integrates an Encode Depthwise Convolutional Network (EDCN) with a Decoder to reduce feature dimensions and personalize models, aligning with the objectives of efficient and tailored FL applications.  \citet{elayan22} introduced a deep FL (DFL) framework aimed at IoT-based healthcare applications, achieving notable success in monitoring accuracy and maintaining data privacy. Their framework is distinctive for splitting the FL process into three phases: initialization, data collection, and model update, streamlining the learning process.

\citet{anand23} explored the effectiveness of FL in a Smart Cities Street Light Monitoring application.  While the FL-based approach did not outperform other machine learning models in terms of accuracy, it significantly enhanced communication efficiency and data privacy in real-world deployments. Furthermore, \citet{fan22} proposed an innovative FL architecture for the Internet of Medical Things (IoMT), named {\em FLDIoMT}, broadening the scope and scalability of IoMT frameworks as well as addressing critical data security concerns. They incorporated data reputation in the global model update process, ensuring that only contributions from local models with significant value are considered.

While various strategies have been proposed to create secure and efficient monitoring systems via FL, it is evident that addressing system vulnerabilities extends beyond ensuring data privacy. Comprehensive security measures are essential to safeguard the normal operation of such systems, underscoring the complexity and multifaceted nature of challenges in FL deployment.

\subsection{FL-based Smart Farms}

\citet{idoje2023federated} utilized FL for crop classification in smart farms, using climatic features and crop types as labels, employing a hyper-tuned Federated Averaging algorithm for precise predictions within a decentralized network, validated through a PySyft, Pytorch, and Syft Testbed. \citet{siniosoglou2023applying} implemented a Centralized FL System (CFLS) for time series forecasting in smart agriculture, using  Long Short-Term Memory (LSTM) models and FL for data privacy and efficient learning with extensive data. \citet{manoj2022federated} explored FL in agricultural risk management for crop yield predictions, showing its superiority in accuracy and privacy over centralized methods. \citet{vimalajeewa2021service} introduced a service-based FL model to predict milk quality, proposing sequential and parallel aggregation methods for enhanced efficiency. While these studies highlight FL's potential in smart agriculture, they focus on non-living targets and lack real-world dataset evaluation. Additionally, they primarily rely on federated averaging, which may be insufficient for complex scenarios -- a gap our work aims to address.

\citet{friha2022felids} introduced FELIDS, a FL-based intrusion detection system (IDS) for securing agricultural IoT systems. FELIDS ensures data privacy through local model updates without sharing raw data, achieving superior detection capabilities as evidenced by datasets like CSE-CIC-IDS2018, MQTTset, and InSDN. On the security enhancement front, \citet{praharaj2023hierarchical} proposed a hierarchical federated transfer learning framework to safeguard cooperative, smart farming against cyber threats by detecting anomalies and malicious activities. However, the absence of empirical evidence in their study leaves the effectiveness and comparative advantage of their framework unverified.

To our knowledge, research on addressing comprehensive security issues in FL-based monitoring systems for smart environments remains scant. Further exploration is required in the FL-based smart farming sector. Our pioneering work introduces an FL-based system for predicting animal diseases, marking the first instance of utilizing FL to oversee animal health conditions in smart farming environments. Moreover, our work diverges from existing energy-efficient FL methodologies by focusing on delivering secure and robust services. Specifically, we concentrate on energy-aware FL strategies tailored for resource-limited, solar sensor-equipped smart farm systems, bridging a critical gap in the literature.

\section{Problem Statement}
\label{sec:problem-statement}
\begin{figure}[t]
\centering
\includegraphics[width=0.5\textwidth]{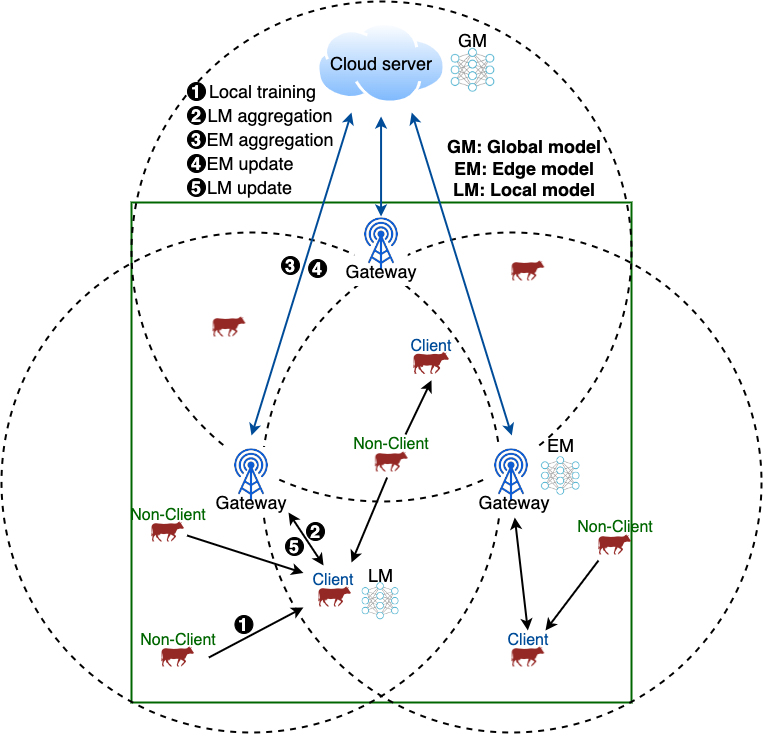}
\caption{Hierarchical FL-based network architecture designed for a wireless solar sensor-based smart farm}
\label{fig:NM}
\vspace{-3mm}
\end{figure}

Our smart farm system employs FL to accurately predict animal disease risks while prolonging system longevity. We conceptualize this system as a {\em hierarchical FL structure}, as depicted in Figure~\ref{fig:NM}. This structure features a global model hosted on a cloud server, with local models operating on LoRa gateways (termed as {\em edge devices}) and Raspberry Pis mounted on animals (referred to as clients). For simplicity, models on gateways are designated as {\em edge models}, and those on clients as {\em local models}, with the central server running the {\em global model}.

Our primary focus lies on the edge level, where gateways execute global models and clients manage local models. Each gateway communicates with a specific set of sensor clients within its communication range. Clients update their models by transmitting them to the gateway offering the highest utility. Detailed explanations on client utility estimation are provided in Section~\ref{subsec:clients-utility-estimation}.

Each gateway in our system seeks to enhance the performance of its edge model by aggregating learning parameters from local models of a carefully chosen set of clients. This selection process prioritizes security, energy efficiency, and fairness. Conversely, clients strive to conserve energy to prolong their operational lifespan while supplying essential updates to improve the edge model's effectiveness. Collectively, our system tackles multi-objective optimization (MOO) challenges through a $\varepsilon$-constraint method, embodied in a scalarization-based MOO function~\cite{cho2017survey}, aiming to:
\begin{equation}
\begin{gathered}
\text{maximize} \quad \mathcal{ACC}(M(s^*)) \\
\text{subject to $\mathcal{EC}(s^*) \geq \varepsilon$}.
\end{gathered}
\label{eq:OF}
\end{equation}
In our system, $M(s^*)$ denotes the edge model trained using a selected set of sensor nodes $s^*$ for the FL aggregation process.  The $\mathcal{ACC}(M(s^*))$ measures the prediction accuracy of $M(s^*)$, and $\mathcal{EC}(s^*)$ quantifies the energy consumption of these selected sensor nodes. Both accuracy $\mathcal{ACC}(M)$ and the target accuracy threshold $\varepsilon$ are normalized within the range $[0,1]$. Our objective is to minimize $\mathcal{EC}(s^*)$ while ensuring that $\mathcal{ACC}(M(s^*))$ meets or exceeds $\varepsilon$. This approach underlines our commitment to developing a sustainable smart farm through hierarchical FL, namely {\tt SusFL}, representing {\em sustainable FL}. 

{\tt SusFL} incorporates a mechanism design-based client selection for FL aggregation, designed to withstand adversarial attacks (detailed in Section~\ref{subsec:threat-model}). We will elaborate on the system's design and its components aimed at achieving sustainability in Section~\ref{sec:proposed-approach}.

\section{System model} \label{sec:system-model}
In this section, we describe the considered network, node, and threat models.

\subsection{Network Model} \label{subsec:network-model}

Our smart farm system employs a network model that integrates solar-powered sensors, wearable Raspberry Pis (R-Pis), Long-Range (LoRa) gateways, and a cloud server, as illustrated in Figure~\ref{fig:NM}. Each animal, such as a cow, is fitted with solar sensors on their ears to monitor body conditions, with the data transmitted to nearby R-Pis. A selected group of animals equipped with R-Pis act as computational clients, processing their data and that from others to train local models—constituting step 1. LoRa gateways, equipped with edge servers, receive these local updates, refine the edge model, and forward the refined model parameters to the cloud server, which then updates the global model in steps 2 and 3. Our approach, {\tt susFL}, focuses on optimizing these processes, particularly for energy-limited devices (i.e., clients). In step 4, the cloud server dispatches the latest model parameters back to the gateways, distributing them to the clients within range in step 5. This configuration capitalizes on LoRa technology to boost IoT connectivity, minimizing costs while extending the range.

A deep learning (DL) model, deployed across local servers on client devices, edge servers, and the central server, is designed to assess the risk of mastitis in animals. It outputs a binary classification: 0 indicates a healthy cow, while 1 signifies a cow diagnosed with mastitis.

\subsection{Node Model} \label{subsec:node-model}

In the smart farm network, sensors periodically transmit data to nearby clients, enabling the training of local models with freshly sensed data. Given these sensors are solar-powered, their energy levels naturally fluctuate due to various environmental influences, including the animals' locations, weather conditions, and seasonal variations in sunlight exposure. Additionally, designing FL systems faces significant challenges, such as high communication and computational costs, along with the need to ensure data privacy. Consequently, the system needs to be resilient, capable of adapting to the dynamic and energy-variable environment of the smart farm, and fortified against potential adversarial attacks.

In our smart farm network, sensors use Bluetooth Low Energy (BLE) to send data to proximal clients, and these clients, in turn, forward their local updates to LoRa gateways. This setup employs the LoRa protocol for long-range communications, effectively covering distances between 5 to 15 $km$ with a data transfer speed of 27 $kbps$. For shorter distances, up to 100 meters, the BLE protocol is used, boasting a faster transfer speed of 2 $Mbps$. 

For energy consumption, the LoRa radio of SAM R34/35 expends about 170 $mW$ during data transmission, while the BLE radio has a lower consumption rate of approximately 11 $mW$. A Raspberry Pi's power usage is 0.117 $W$ per second when idle, increasing to 0.172 $W$ per second under load. Sensor nodes, once fully charged, hold an initial energy reserve of 5 $kW$. The charging efficiency for solar-powered sensors varies with light exposure—about 10 $mW/cm^2$ in outdoor settings and 0.1 $mW/cm^2$ indoors.

We distinguish between two sensor node types in our analysis: 
\begin{itemize}
\item {\em Normal sensor node}~\cite{manualR3}: This node lacks the computational resources for local model training, instead periodically sending its data to a Raspberry Pi (R-pi)-based node via BLE.
\item {\em R-pi-based sensor node}~\cite{caria2017smart, manualR2, rasp2023}: This node is capable of gathering data from normal nodes within its range and training local models. This node decides on its participation in the FL aggregation by sending its local model parameters to the edge model, thereby acting as a client within the FL framework.
\end{itemize}

\subsection{Threat Model} \label{subsec:threat-model}
To understand the vulnerabilities within FL systems, we examine the following types of adversarial attacks:
\begin{itemize}
\item {\em Byzantine attacks} disrupt the FL training process by injecting arbitrary metrics via Stochastic Gradient Descent (SGD) updates~\cite{blanchard17}. These attacks primarily target local devices or clients, prolonging their learning duration or leading to model divergence.

\item {\em Backdoor attacks} compromise the integrity of edge and global models through malicious clients that submit altered local model updates~\cite{eug21}. The objective of backdoor attackers is to preserve high prediction accuracy during testing to evade detection while causing the model to incorrectly classify a specific target class.

\item {\em Collaborative attacks} involve multiple compromised clients working together to degrade the global model's accuracy~\cite{bag20}. This type of attack affects both the global model on the central server and the edge models on gateways. Attackers may adjust training hyperparameters or alter model weights before these are sent to the edge model. The success rate of backdoor attacks increases with the proportion of attacker-controlled clients, surpassing the effectiveness of conventional data poisoning strategies.
\end{itemize}

To assess the impact of these attacks, we analyze the attack success probability, denoted as $P_A$, which represents the likelihood of an attacker successfully executing an attack at any given time $t$. Our primary goal is to create a sustainable FL-based monitoring system characterized by its resilience and capability of functioning effectively under adversarial attacks, including Byzantine, backdoor, and collaborative attacks. Consequently, our work does not concentrate on developing specific defenses against these attacks. Instead, we emphasize the selection of trustworthy clients and the secure aggregation of local model parameters to ensure the system's tolerance and robustness against such cybersecurity threats.

\section{Proposed Approach: {\tt SusFL}} \label{sec:proposed-approach}

In the given network, we consider a group of $n$ clients, denoted by $N = \{1, \ldots, n\}$, each possessing a local model eligible for selection. The cost of including client $i$'s local model in the aggregation process is represented by $c_i$, a known value publicly. With a total budget constraint of $B$, we aim to ensure that the selected clients maintain adequate energy reserves after completing a given FL task. To achieve this, we propose to develop a client selection mechanism, $\mathcal{M}$, aimed at identifying an optimal subset of clients. This subset will contribute to the FL model aggregation, optimizing the balance between achieving the desired accuracy of the edge model and minimizing energy consumption.

\begin{figure}[h]
\centering
\includegraphics[width=0.48\textwidth]{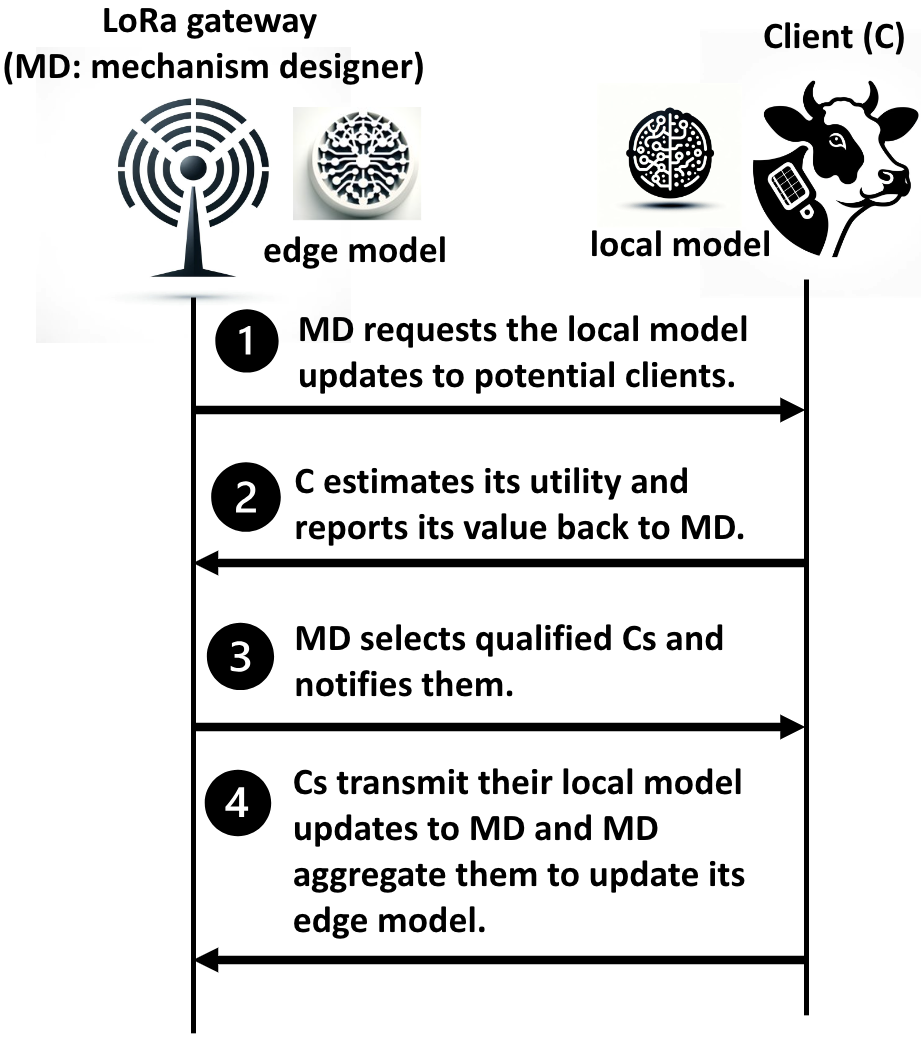}
\caption{Key steps for an edge server (i.e., mechanism designer) to aggregate local model updates from selected clients}
\label{fig:proposed-sys}
\vspace{-3mm}
\end{figure}
Figure~\ref{fig:proposed-sys} describes the FL model aggregation process in four key steps in our proposed {\tt SusFL} system:
\begin{enumerate}
\item {\em Request for local model updates}: Each gateway's server requests local updates from clients within its operational area.

\item {\em Utility estimation and response}: Clients calculate their utility based on the anticipated energy expenditure for participating in the FL process, as defined in Eq.~\eqref{eq:client-utility}. Clients with utilities that are equal to or larger than a predefined threshold $\theta$ communicate their values, as formulated by Eq.~\eqref{eq:value-e}, back to the gateway.

\item {\em Client selection and notification}: The mechanism designer, operating at the gateway, selects clients for the FL process based on their values and notifies the chosen clients.

\item {\em Local update transmission and model aggregation}: Selected clients transmit their local updates to the gateway's edge server, aggregating these updates into an enhanced edge model, focusing on high-quality data as described in Section~\ref{subsec:data-quality-aggregation}. This model is used to assess the health conditions of cows on the considered {\tt SusFL}-based smart farm.
\end{enumerate}
After completing each FL cycle, gateways upload their updated models to the central server, consolidating them into a new global model. This updated global model is then disseminated back to the gateways, ensuring continuous improvement and accuracy of the system's predictive capabilities.  We describe the key design components of the proposed {\tt SusFL} in the following sections.

\subsection{Clients' Utility Estimation} \label{subsec:clients-utility-estimation}

Upon receiving a local model update request from its gateway, a client assesses the task's utility to determine its participation in the aggregation process. The utility of client $i$ upon receiving this request from gateway $j$ at time step $t$, denoted as $u_t(i, j)$, is calculated by:
\begin{equation}
\label{eq:client-utility}
u_t(i, j) = e_t(i) - ec_t(i, j).
\end{equation}
The utility calculation incorporates $e_t(i)$, the current energy level of client $i$, and $ec_t(i, j)$, the expected energy consumption of client $i$ when participating in the FL aggregation process with gateway $j$. Both variables, $e_t(i)$ and $ec_t(i, j)$, are normalized to the range $[0,1]$ as real numbers. The expected energy consumption, $ec_t(i, j)$, is determined by:
\begin{gather}
ec_t(i, j) 
= \underbrace{|D_{i}^{t}| \times r \times E_{\mathrm{R-pis}}}_{\text{for training}} + \underbrace{\mathrm{Dis}(p_{i}^{t}, p_{j}^{t}) \times E_T}_{\text{for transmission}}.,
\end{gather}
The expected energy consumption, $ec_t(i, j)$, is calculated based on $|D_{i}^{t}|$, the volume of data client $i$ uses for local training at time step $t$, and $r$, the rate at which training time extends per additional data sample. It also considers $E_{\mathrm{R-pis}}$, the energy required to process the specified workload per second. The distance between client $i$ and gateway $j$ during the FL process at time step $t$, denoted as $\mathrm{Dis}(p_{i}^{t}, p_{j}^{t})$, and $E_T$, the energy for transmitting local updates via LoRa, are also factored into the calculation.

The utility function enables client $i$ to evaluate the benefit of participating in the aggregation process at time step $t$. If participating would result in the client's energy being entirely depleted from transmitting data to the edge model, leading to a utility of $u_t(i, j) \leq 0$, the client will opt out of the aggregation. This decision adheres to the {\em individual rationality} property within mechanism $\mathcal{M}$, ensuring clients participate only when the utility is positive, thus preserving their normal operation until a more opportune moment arises. Upon opting to contribute, client $i$ communicates its value $v_i$ to the edge server, which then assesses client selection. The value $v_i$ reflects the data quality of client $i$'s model update, as determined by~\cite{kang19}, and formulated by:
\begin{equation}\label{eq:value-e}
v_i = \frac{\Psi}{\log (\frac{1}{\varepsilon_i})},
\end{equation}
The value $v_i$ is influenced by $\Psi$, a coefficient related to the number of local model iterations impacted by local data accuracy, and $\varepsilon$, which denotes the iteration count for a local model update to maintain a constant global accuracy. This assessment is supported by theoretical guarantees based on empirical evaluations rather than hypothetical expectations. Importantly, $v_i$ assumes a negative value if $\Psi > 0$ and $\varepsilon_i > 1$, indicating that the closer $v_i$ is to zero, the higher its perceived value.

To address the Multi-Objective Optimization (MOO) challenge discussed in Section~\ref{sec:problem-statement} and streamline the search space, gateways initially identify clients whose participation in the selection process could ensure a minimum accuracy of $\varepsilon$. Consequently, only clients with a value of $v_i \geq \theta$ qualify as candidates for aggregation. Here, $\theta$ represents the threshold that defines the subset of clients eligible for further consideration in the client selection phase.

\subsection{Client Selection Mechanism}\label{subsec:clients-selection-mechanism}

Mechanism $\mathcal{M}$ is designed to fulfill specific properties, simplifying our discussion by excluding the notation $j$. This approach generalizes the utility of client $i$ across any scenario where it receives a request for a local model update to participate in the FL aggregation of gateway $j$, as depicted in Figure~\ref{fig:proposed-sys}.

\begin{itemize}
\item \textit{Truthfulness}~\cite{bei2012budget}: $\mathcal{M}$ is truthful if each client $i$'s dominant strategy is to report true information when
\begin{equation}
u(i) \geq u(i^{'}),
\label{eq:thruthful}
\end{equation}
where $i^{'}$ is any client's information that $i^{'} \neq i$. 
\item \textit{(Weak) Budget balance}~\cite{chawla2011bayesian}: $\mathcal{M}$ is budget-balanced when 
\begin{equation}
\begin{aligned}
\sum_i^N \mathcal{COM_E}(i) \leq B,
\end{aligned}
\end{equation}
where $\mathcal{COM_E}(i)$ represents the communication cost incurred when integrating client $i$ into the aggregation. LoRa gateways, not being limited by energy constraints for model training, could theoretically accept clients' local updates immediately. However, to minimize energy consumption on the client side and preserve the operational longevity of sensor nodes, it is imperative from a global standpoint to restrict the number of communication rounds. This strategy ensures that clients maintain sufficient energy levels for regular operations.
\item \textit{Individual rationality}~\cite{gresik1991ex}: $\mathcal{M}$ is individual rational if 
\begin{equation}
\forall i \quad u(i) \geq 0.
\label{eq:IR}
\end{equation}
Our utility function is designed to assess the variance between a client's present energy status and the energy expenditure associated with contributing to aggregation $j$. Thus, clients will partake in aggregation $j$ solely if they anticipate sustaining their energy levels post-participation, ensuring no energy depletion occurs from engaging in the task for gateway $j$.
\end{itemize}
The mechanism designer (MD) will aim to maximize its social welfare by selecting an optimal set of clients via mechanism $\mathcal{M}$. This is formulated by:
\begin{equation}
\begin{aligned}
\text{maximize} \quad \sum_i^n u(i)  \\
\text{satisfy properties (\ref{eq:thruthful}) - (\ref{eq:IR})}
\end{aligned}
\label{eq:social-welfare}
\end{equation}
To fulfill this goal, the defined properties ensure that each client can engage in the aggregation process without compromising their energy reserves. Moreover, the design of the client's utility function is independent of the information disclosed to the gateway, such as data quality. This design choice ensures that clients have no incentive to misreport their values, as doing so will not adversely affect their utility.

Given the significant resource constraints, such as energy, implementing mechanism $\mathcal{M}$ with efficiency is paramount. This challenge can be modeled as a 0/1 knapsack problem, addressed through the dynamic programming approach~\cite{toth1980dynamic}. This method guarantees an optimal solution. The time complexity for the dynamic programming solution is $O(N \times W)$, where $N$ is the number of clients and $W$ represents the gateway's budget constraints. Dynamic programming resolves complex problems by dividing them into smaller, overlapping subproblems, addressing each once. For each of the $W$ possible budgets, we determine the optimal value for every client, resulting in $N \times W$ subproblems. Each subproblem is solvable in constant time, or $O(1)$.

\subsection{Quality-Aware Parameter Aggregation} \label{subsec:data-quality-aggregation}
Upon receiving local updates from the chosen clients, gateways proceed to aggregate these updates to train new edge models. To achieve high model accuracy, the strategy employs FedAvg~\cite{shi2022fedfaim}, focusing on utilizing high-quality data. This approach involves a weighted parameter aggregation, where the weighting is determined by the reported data quality of each client. The formulation of this process is given by:
\begin{equation}
w_{new} = \frac{\sum_i^k \frac{v_i}{v_{max} - v_{min}} w_i}{k}. 
\end{equation}
The weighting of each local parameter $w_i$ is determined by $\frac{v_i}{v_{max}-v_{min}}$, positioning the data quality within a $[0,1]$ scale in the overall distribution, with $k$ representing the total number of participating clients. Aggregation occurs both at the edge server level, involving gateway and client updates, and at the cloud server level, with updates from the edge. This method capitalizes on the data quality values reported by clients to streamline the process. Following aggregation, gateways forward the updated edge model parameters to the central server, where edge updates are compiled to construct the global model. During training, local devices employ a loss function to assess model performance on their datasets, with gradients informing local updates. The effectiveness of our {\tt SusFL} system is gauged through the global model's performance, as discussed in Section~\ref{sec:results-analyses}.

\section{Experimental Setup} \label{sec:exp-setup}

\begin{table}[t]
\centering
\caption{\sc Dataset Description}
\label{tab:mastitis_data}
\begin{tabular}{|P{2.5cm}|P{5cm}|}
        \hline
        Metric & Description \\
        \hline
        Serial & A unique animal identifier \\
        \hline
        Size-udder & Size of an udder (udder front left, front right, rear left, and rear right) for inhale and exhale limit \\
        \hline 
        Average temperature & Average body temperature in Celsius \\
        \hline
        Hardness & Hardness of an udder \\
        \hline
        Pain-level & Pain due to swelling of an udder \\
        \hline
        Average-activity & Average activity recorded by the number of steps taken \\
        \hline
        Battery-level & Residual battery life \\
        \hline 
        Timestamp & Date and time of transmission \\
        \hline 
    \end{tabular}
\end{table}

\subsection{Parameterization} \label{subsec:datasets}

This study leverages clinical mastitis data in cows, captured via IoT sensors on the udder, to detect the disease~\cite{ankitha2020data}. The dataset consists of 6,600 entries, with three records per cow, featuring 15 attributes monitored by flex and temperature sensors connected to Raspberry Pis and digital-to-analog converters. Additionally, we utilize data from Virginia Tech's SmartFarm Innovation Network (TM), a project of the College of Agriculture and Life Sciences, which records cow movement activities. This network serves as a hub for collecting and analyzing data across Virginia farms, indicating cows typically move at speeds within the range of $[1, 2]$ meters per second. Movement probability for cow $i$, denoted as $P_{mv}^i$, is modeled by a normal distribution with an average speed of 1.5 $m/s$ and a standard deviation of 0.1 $m/s$. Our system, detailed in Table~\ref{tab:mastitis_data}, applies FL to diagnose animal diseases using this data alongside information from sensors. Additionally, we crafted semi-synthetic datasets by introducing simulated threats to the original data, following the threat models in Section~\ref{subsec:threat-model}.

This work encompasses a farm spanning 40 acres, approximately 160,000 square meters, with each side measuring 400 meters. It focuses on monitoring 30 cows using three gateways to ensure efficient surveillance over the 48-hour simulation period. Each gateway implements an edge model, leveraging our {\tt SusFL} mechanism to predict animal diseases, aiming for optimal system performance given the current conditions. We classify 60\% of the cows as clients equipped with Raspberry Pi (R-pi)-based sensor nodes, with the remaining serving as standard sensor nodes, as described in Section~\ref{subsec:node-model}. Gateways solicit updates from these client nodes at 60-minute intervals, designated as $T_u$. All sensor nodes start with a random initial energy level, $E_{init}$, within the range of $[0.3, 0.8)$. Table~\ref{tab:param-default} presents the key design parameters, their meanings, and default values. Initially, we posit that 30\% of the sensors, denoted as $P_C$, are compromised at the system's onset. The model assumes full trust in both gateways and the cloud server, with attackers solely targeting sensor nodes.

\begin{table}[t]
\centering
\caption{\sc {\centering Key Design Parameters, Meanings, \& Default Values}}
\label{tab:param-default}
\begin{tabular}{|P{1.1cm}|P{5cm}|P{1.1cm}|}
\hline
Notation & Meaning & Value \\
\hline
$n$ & Total number of sensors(cows) & 30\\
\hline
$N$ & Total number of clients & 20\\
\hline
$P_{mv}^1$ & Probability of cow $i$ to move & [0.3,0.7]\\
\hline 
$P_{A}$ & Probability for an attacker or a compromised node to perform a certain attack (e.g., ) & $0.1$ \\
\hline
$P_{C}$ & Percentage of compromised clients in sensor network & $0.3$ \\
\hline
$T_s$ & Time interval for a sensor to send sensed data & 30 s\\
\hline
$T_u$ & Time interval for a gateway to request local updates & 1 hr\\
\hline
$T_g$ & Time interval for a gateway to report edge models to central server & 1 hr\\
\hline
$E_{init}$ & Initial energy level of sensors  & $[0.3,0.8)$ \\ 
\hline
$\varepsilon$ & Threshold for minimum energy level & 0.15 \\
\hline
$B$ & Number of communication rounds budget for each gateway & 5 \\
\hline
\end{tabular}
\vspace{-3mm}
\end{table}

\subsection{Metrics} \label{subsec:metrics}
We use the following metrics for performance validation:
\begin{itemize}
\item {\bf Prediction accuracy} evaluates the accuracy of the global model's predictions on animal diseases against actual outcomes. Specifically, it assesses the system's ability to correctly identify cases of mastitis in cows, calculated as the proportion of accurate predictions out of the total predictions made throughout the simulation.
\item {\bf Energy consumption} ($\mathcal{EC}$) quantifies the overall energy usage by the system for communications and computations within the FL framework. The $\mathcal{EC}$ is formulated by:
\begin{equation}
\label{eq:energy-consumption-metric}
\mathcal{EC} = \mathcal{COM_E} + \mathcal{COMP_E},
\end{equation}
where $\mathcal{EC}$, the total energy consumption, comprises two main components: $\mathcal{COM_E}$, the energy used for communications between clients and gateways, and $\mathcal{COMP_E}$, the cumulative energy expended by sensor nodes for system operations, including model training and energy depletion over time. The $\mathcal{COM_E}$ is calculated by:
\begin{equation}
\mathcal{COM}_E = \sum_{i=0}^l \mathcal{E}_{CR}^{i},
\end{equation}
where $\mathcal{E}_{CR}^{i}$ represents the energy consumed in a single communication round between a client and a gateway, and $l$ signifies the total number of communication rounds throughout the simulation. The $\mathcal{COMP_E}$ accounts for the energy utilized in model training and the natural energy drain experienced during the system's operational period.  The $\mathcal{COMP_E}$ is given by:
\begin{gather}
\label{eq:re-at}
\mathcal{COMP_E} = \mathcal{E}_{TC} + \mathcal{E}_{\mathrm{active}} +\mathcal{E}_{\mathrm{sleep}} \\ 
= \frac{|S^*|e_{TC}}{E_S} +  \frac{T_u}{E_S} (d_{\mathrm{active}} +d_{\mathrm{sleep}}), \nonumber 
\end{gather}
where $\mathcal{COMP_E}$ accounts for the energy dynamics of the participating client set $|S^*|$, including $e_{TC}$, the energy a client consumes to join the aggregation process, and $E_S$, the full charge energy level of a sensor. It further considers $d_{\mathrm{active}}$ and $d_{\mathrm{sleep}}$, the energy depletion rates per second in active and sleep modes, respectively.  $T_{u}$ refers to the interval at which local updates are requested by the edge server. This model focuses solely on the energy consumption of sensor nodes, excluding the energy transactions between gateways and the central server, as they do not face energy limitations.
\item {\bf Mean Time Between Failures (MTBF)}~\cite{colombo1990systems, cho2019stram} quantifies the average duration of system operation without interruptions, calculated as:
\begin{equation}
MTBF = \frac{\sum_{f \in F} (u_{s,f} - d_{s,f})}{|F|},
\end{equation}
where $d_{s,f}$ marks the commencement of downtime, $u_{s,f}$ the onset of uptime, and $F$ the collection of failure instances. A system is considered to have failed when the average energy level across all nodes falls below $\varepsilon$. 

\item {\bf Social Welfare} encapsulates the collective utility of all sensor clients, computed as detailed in Eq.~\eqref{eq:social-welfare}.
\end{itemize}
\subsection{Comparing Schemes} \label{subsec:comparing-schemes}
We compare the proposed {\tt SusFL} against the following state-of-the-art (SOTA) FL schemes to evaluate its effectiveness:
\begin{itemize}
\item {\bf FedAvg}~\cite{t2020personalized} employs a variant of the Stochastic Gradient Descent (SGD) where clients independently execute SGD. The server then derives global model parameters by averaging these local models, bypassing the need to aggregate the entire dataset.

\item {\bf FedProx}~\cite{li2020federated} modifies FedAvg to better handle client heterogeneity in computational resources. It incorporates a proximal term into the local optimization problems, enabling variable local updates and allowing for the management of statistical heterogeneity across clients.

\item {\bf FLTrust}~\cite{zhang2020enabling} enhances robustness against data poisoning and backdoor attacks, minimizing performance impact. It utilizes trusted execution environments to periodically verify a subset of clients, ensuring integrity.

\item {\bf DivFL}~\cite{balakrishnan2022diverse} promotes communication efficiency by introducing a diverse client selection mechanism. Through a greedy algorithm, it selects an optimal subset of clients for each aggregation round, aiming to diversify the gradient space and hasten training.

\item {\bf GreenFL}~\cite{Kim2022GreenQF} introduces a green-quantized FL approach that incorporates stochastic quantization in both local training and data transmission. Clients maintain a Quantized Neural Network (QNN) with adjustable precision levels, addressing the energy-accuracy trade-off through multi-objective optimization of precision settings.
\end{itemize}
\section{Results and Analyses} \label{sec:results-analyses}

To assess the effectiveness of the proposed {\tt SusFL} scheme alongside the five existing FL schemes, we conducted 50 simulation runs using the parameter configurations in Section~\ref{sec:exp-setup}. The results presented for each scheme are the average outcomes derived from these 50 simulations, ensuring valid evaluation and comparison.

\subsection{Comparative Performance Analyses}

\begin{figure*}[!ht]
  \centering
  \subfigure{
    \includegraphics[width=0.7\textwidth, height=0.027\textwidth]{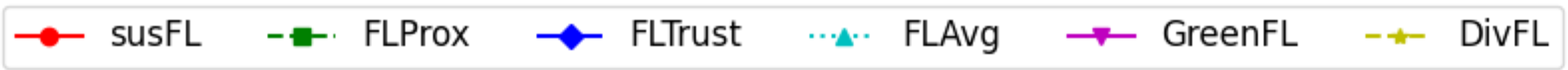}}
   \setcounter{subfigure}{0}
    \vspace{-3mm}
    
  \subfigure[Prediction Accuracy]{
    \includegraphics[width=0.23\textwidth, height=0.2\textwidth]{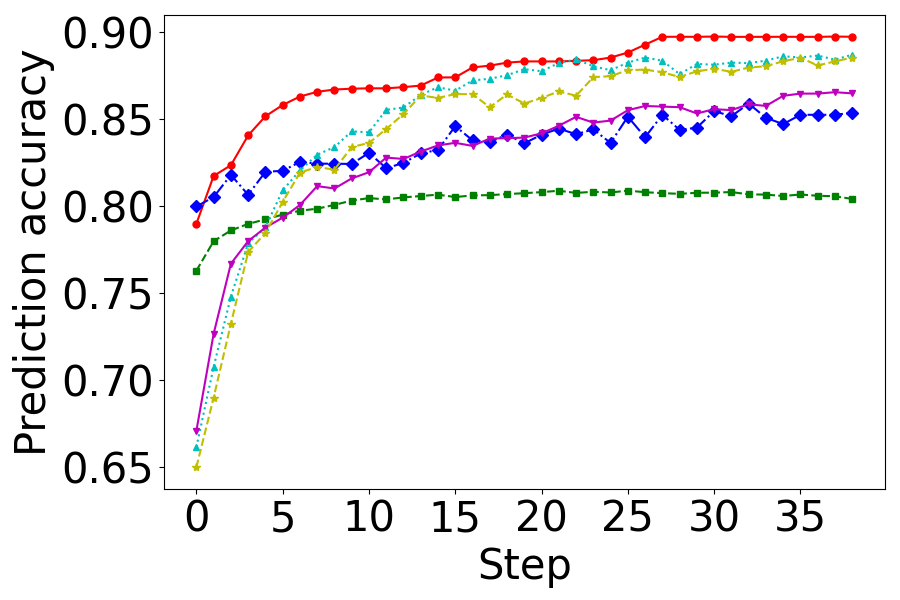}}
  \subfigure[Energy Consumption]{
    \includegraphics[width=0.23\textwidth, height=0.2\textwidth]{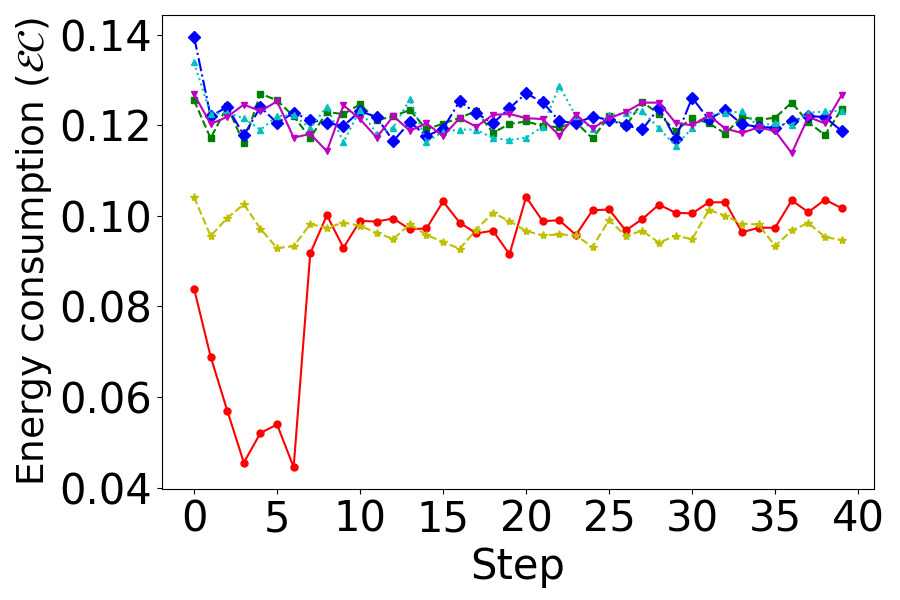}}
  \subfigure[Social Welfare]{
    \includegraphics[width=0.23\textwidth, height=0.2\textwidth]{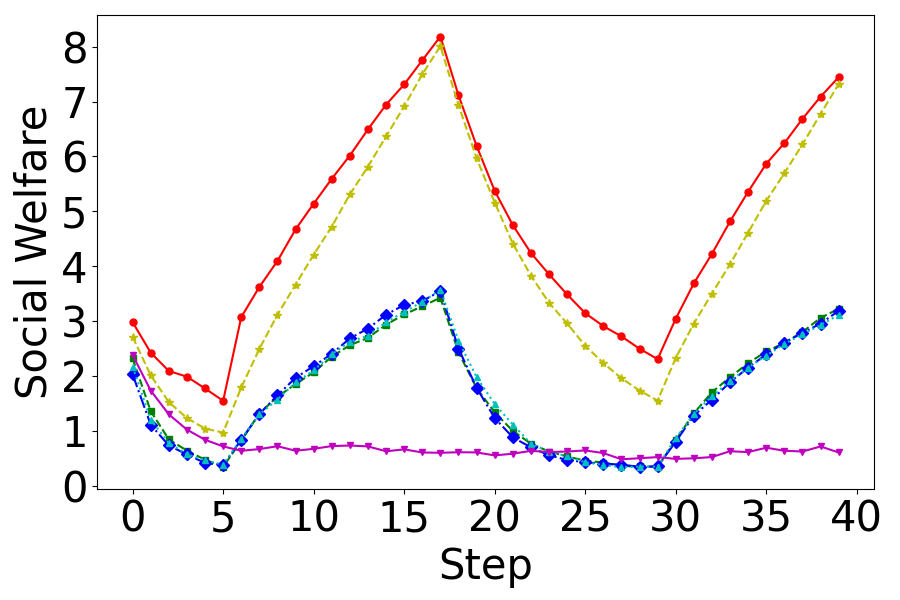}}
    \subfigure[MTBF]{
    \includegraphics[width=0.23\textwidth, height=0.2\textwidth]{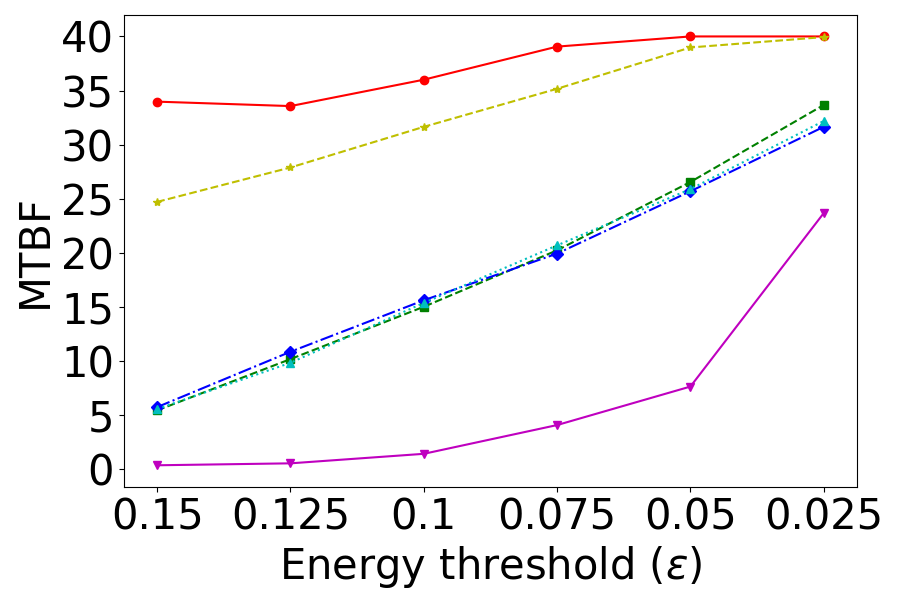}}
    \caption{Comparative performance analysis during training time}
\label{fig:cp}
\vspace{-2mm}
\end{figure*}

Figure~\ref{fig:cp} showcases the FL training progress across six schemes, as detailed in Section~\ref{subsec:comparing-schemes}, including our proposed {\tt SusFL} plus the five SOTA schemes. Our {\tt SusFL} surpasses the five other schemes in prediction accuracy (Figure~\ref{fig:cp}(a)), energy consumption (Figure~\ref{fig:cp}(b)), social welfare (Figure~\ref{fig:cp}(c)), and mean time between failures (MTBF) (Figure~\ref{fig:cp}(d)). This superior performance underscores the efficiency of {\tt SusFL}'s quality-aware client selection strategy, which excludes clients with poor-quality data, thereby enhancing prediction accuracy. Additionally, by allowing clients to assess their utility for participation, {\tt SusFL} achieves minimal energy usage throughout the simulation, leading to the highest social welfare and MTBF. Notably, {\tt SusFL}'s energy consumption is particularly low at the simulation's start, attributed to the initial absence of sunlight for charging sensors, indicating a strategic energy conservation for FL operations. The fluctuation in social welfare (Figure~\ref{fig:cp}(c)) mirrors the solar pattern, highlighting the dynamic nature of the smart farm environment. Among the SOTA schemes, {\tt DivFL} demonstrates the lowest energy usage due to its efficient client selection method that bolsters learning effectiveness. Conversely, {\tt GreenFL} exhibits the highest energy demand, attributed to the added computational load from managing a QNN. These findings emphasize the importance of minimizing computational demands on energy-limited devices in resource-constrained FL systems for smart environments.

\subsection{Sensitivity Analyses}

\subsubsection{\bf Effect of Varying Attack Severity ($P_A$)}

\begin{figure*}[!ht]
  \centering
  \subfigure{
    \includegraphics[width=0.7\textwidth, height=0.027\textwidth]{figs/legend.png}}
    
   \setcounter{subfigure}{0}
    \vspace{-3mm}
    
  \subfigure[Prediction Accuracy]{
    \includegraphics[width=0.23\textwidth, height=0.2\textwidth]{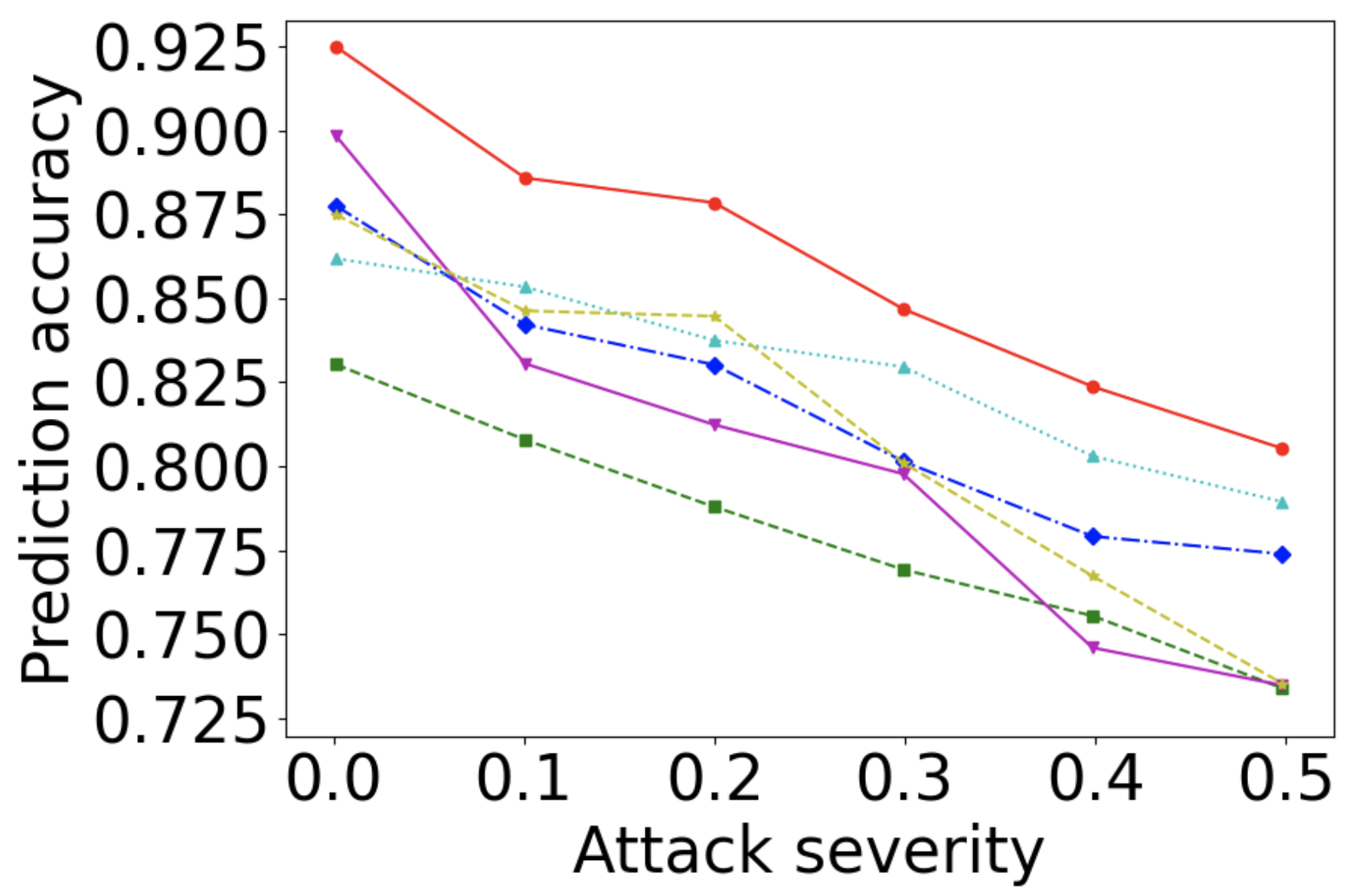}}
  \subfigure[Energy Consumption]{
    \includegraphics[width=0.23\textwidth, height=0.2\textwidth]{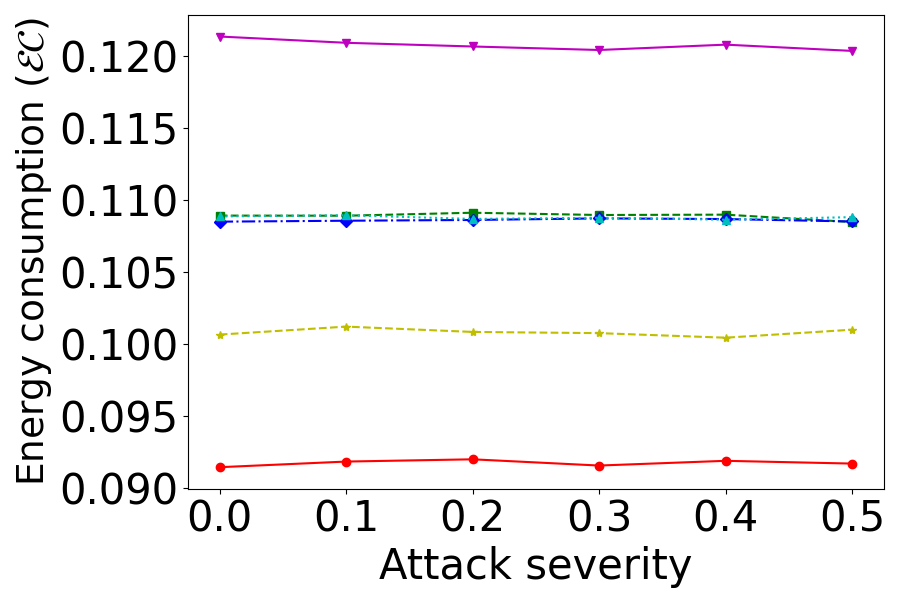}}
  \subfigure[Social Welfare]{
    \includegraphics[width=0.23\textwidth, height=0.2\textwidth]{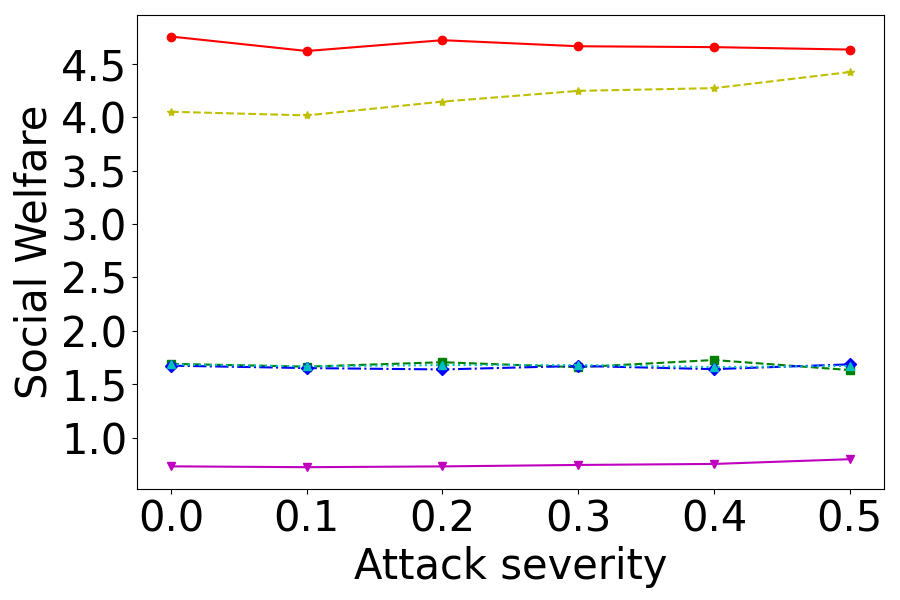}}
    \subfigure[MTBF]{
    \includegraphics[width=0.23\textwidth, height=0.2\textwidth]{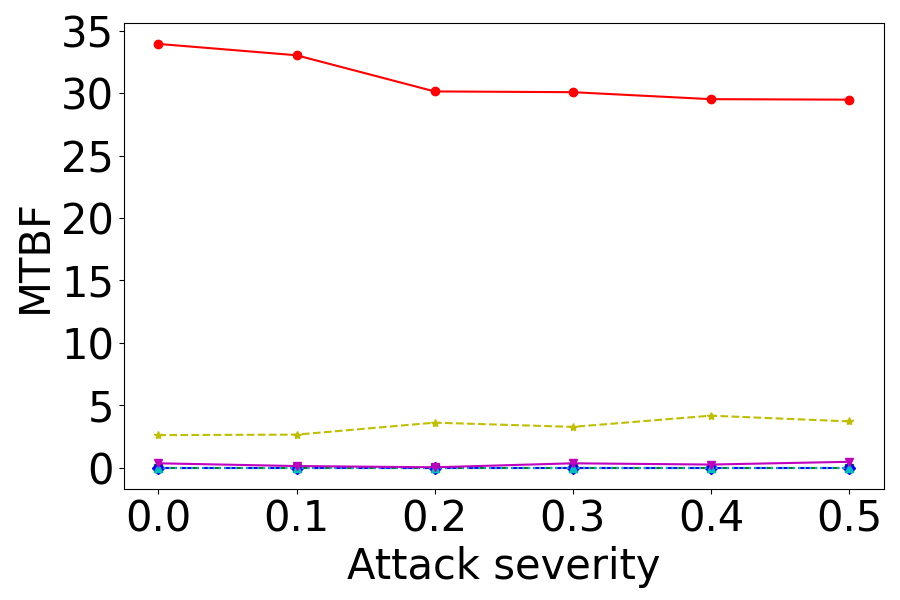}}
    \caption{Effect of Varying Attack Severity ($P_A$)}
\label{fig:effect-attack}
\vspace{-2mm}
\end{figure*}

Figure~\ref{fig:effect-attack} illustrates the impact of different levels of attack severity ($P_A$) on the performance metrics of FL schemes. An increase in $P_A$ results in lower prediction accuracy due to the inclusion of more compromised sensors, undermining the model's performance. However, attack severity has minimal effect on energy metrics, such as energy consumption and MTBF, since the adversarial strategies do not aim to exhaust client energy or reduce their operational lifespan.

For {\tt SusFL}, increased attack severity leads to a decline in social welfare and MTBF as illustrated in Figure~\ref{fig:effect-attack}(c) and (d). This decline necessitates greater client involvement in FL tasks to uphold prediction accuracy. Although this situation slightly raises energy consumption (Figure~\ref{fig:effect-attack}(b)), the increase is marginal when compared to the gap between {\tt SusFL} and other SOTA schemes, rendering the consumption curve nearly flat. Overall, {\tt SusFL} remains superior to its peers, such as {\tt DivFL} and {\tt FLProx}, in handling varying $P_A$, showcasing robust performance across metrics.  

\subsubsection{\bf Effect of Node Density}

\begin{figure*}[!ht]
  \centering
  \subfigure{
    \includegraphics[width=0.7\textwidth, height=0.027\textwidth]{figs/legend.png}}
    
   \setcounter{subfigure}{0}
    \vspace{-3mm}
    
  \subfigure[Prediction Accuracy]{
    \includegraphics[width=0.23\textwidth, height=0.2\textwidth]{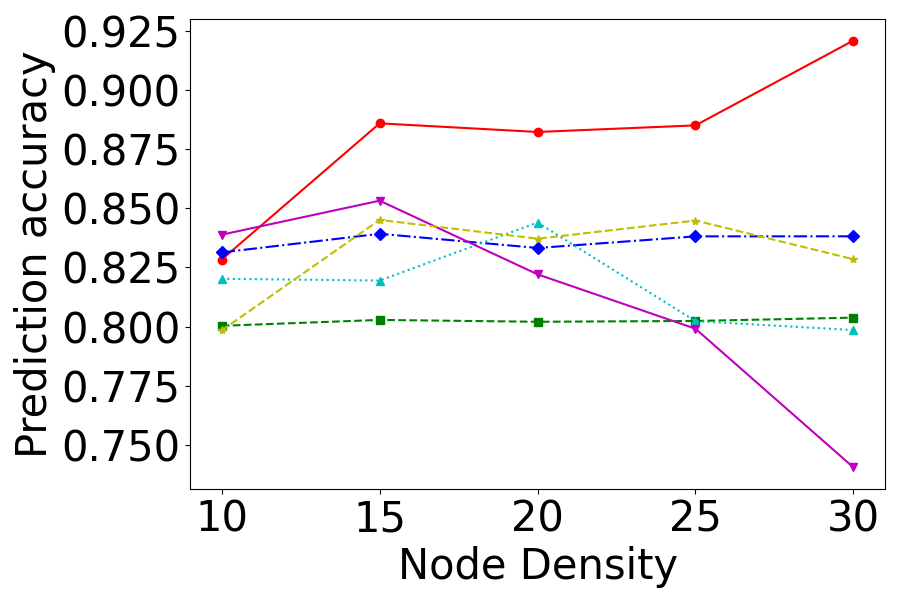}}
  \subfigure[Energy Consumption]{
    \includegraphics[width=0.23\textwidth, height=0.2\textwidth]{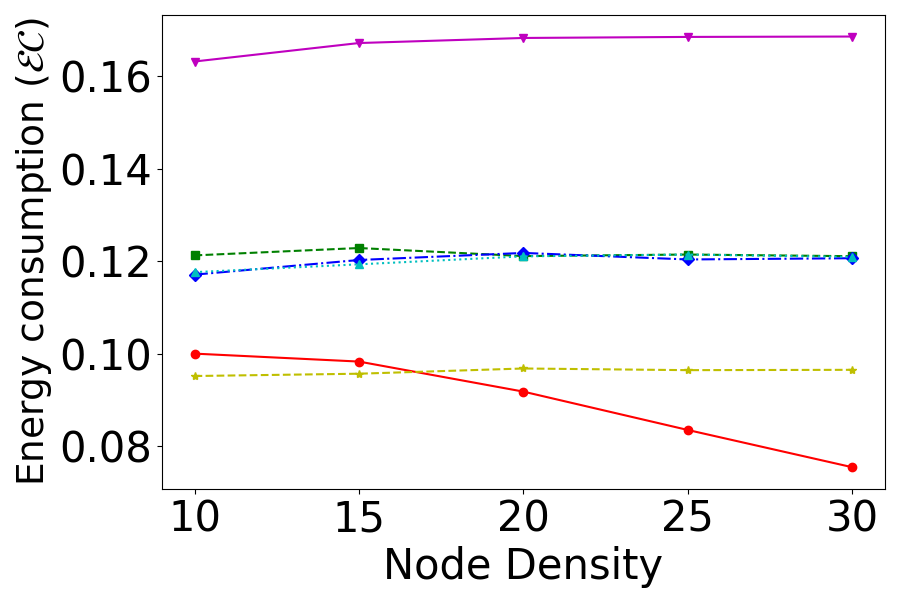}}
  \subfigure[Social Welfare]{
    \includegraphics[width=0.23\textwidth, height=0.2\textwidth]{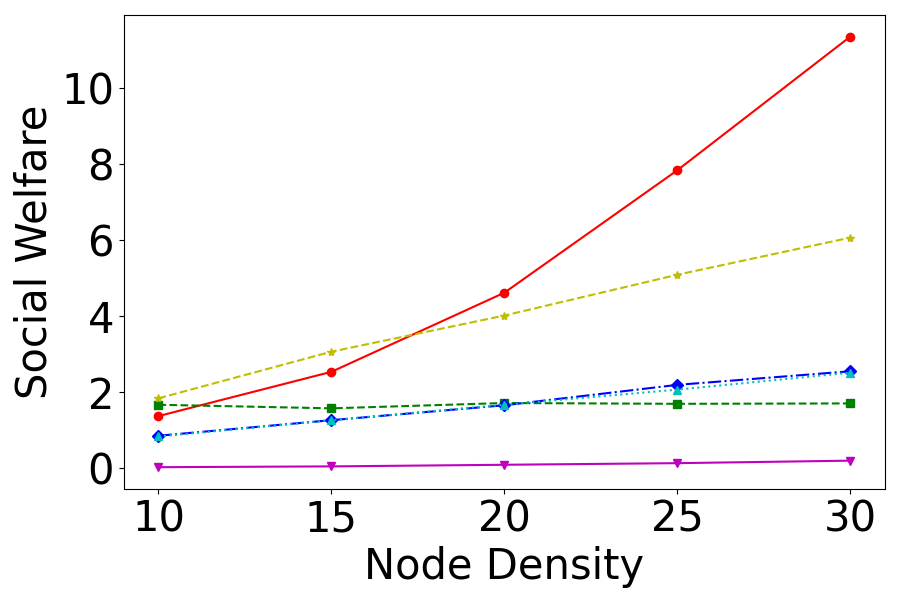}}
    \subfigure[MTBF]{
    \includegraphics[width=0.23\textwidth, height=0.2\textwidth]{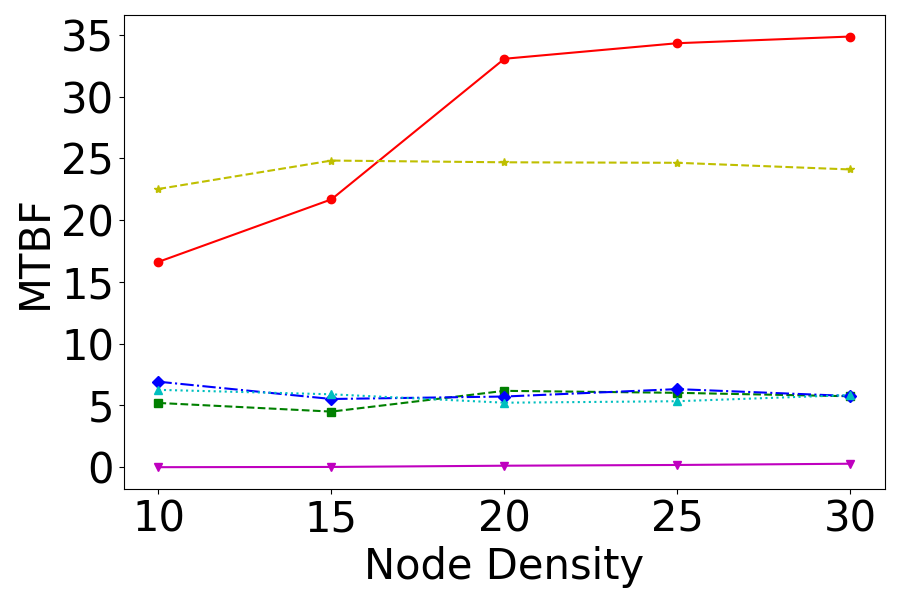}}
    \caption{Effect of Varying Node Density}
\label{fig:effect-node}
\vspace{-2mm}
\end{figure*}

Figure~\ref{fig:effect-node} investigates the performance metrics' response to the varying number of client sensor nodes within the network. The findings demonstrate {\tt SusFL}'s adeptness at managing FL operations across expansive sensor networks. A notable observation is the improvement in prediction accuracy with {\tt SusFL}, attributed to a broader distribution of clients across the farm. This dispersion increases the likelihood of each client possessing high-quality data, thereby providing gateways with more selection options to maintain high accuracy. Consequently, this strategic client selection leads to reduced average energy consumption (Figure~\ref{fig:effect-node}(b)), enhancing social welfare (Figure~\ref{fig:effect-node}(c)) and MTBF (Figure~\ref{fig:effect-node}(d)).

Contrastingly, the performance of other schemes appears relatively unaffected by changes in node density across all metrics, with the exception of {\tt GreenFL}'s prediction accuracy. As the number of clients grows, {\tt GreenFL}'s prediction accuracy diminishes, suggesting a potential shortfall in managing clients with suboptimal local models. This highlights {\tt SusFL}'s superiority in optimizing both the quality and efficiency of FL operations in large-scale sensor networks.  

\subsubsection{\bf Effect of Initial Energy Levels ($E_{init}$) on clients}

\begin{figure*}[!ht]
  \centering
  \subfigure{
    \includegraphics[width=0.7\textwidth, height=0.027\textwidth]{figs/legend.png}}
    
   \setcounter{subfigure}{0}
    \vspace{-3mm}
    
  \subfigure[Prediction Accuracy]{
    \includegraphics[width=0.23\textwidth, height=0.2\textwidth]{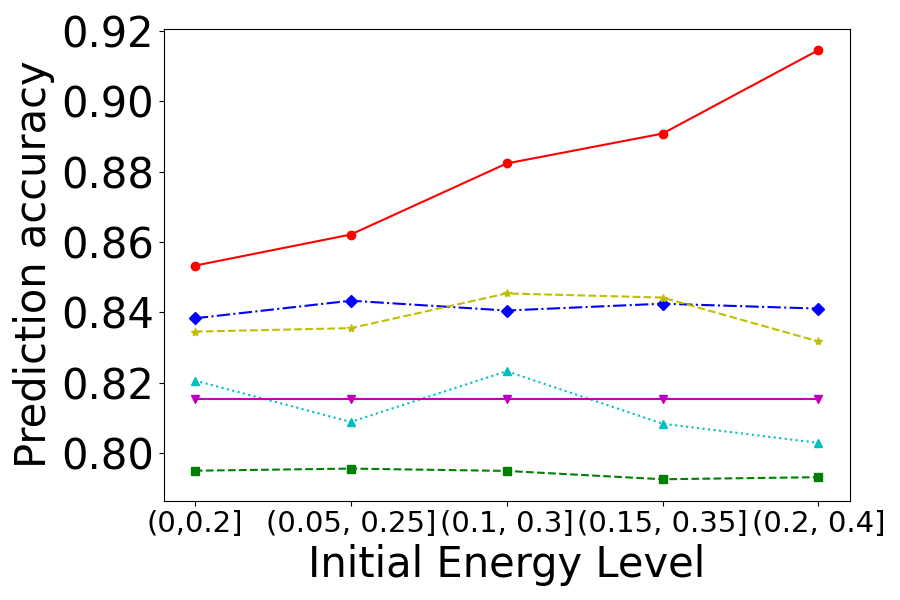}}
  \subfigure[Energy Consumption]{
    \includegraphics[width=0.23\textwidth, height=0.2\textwidth]{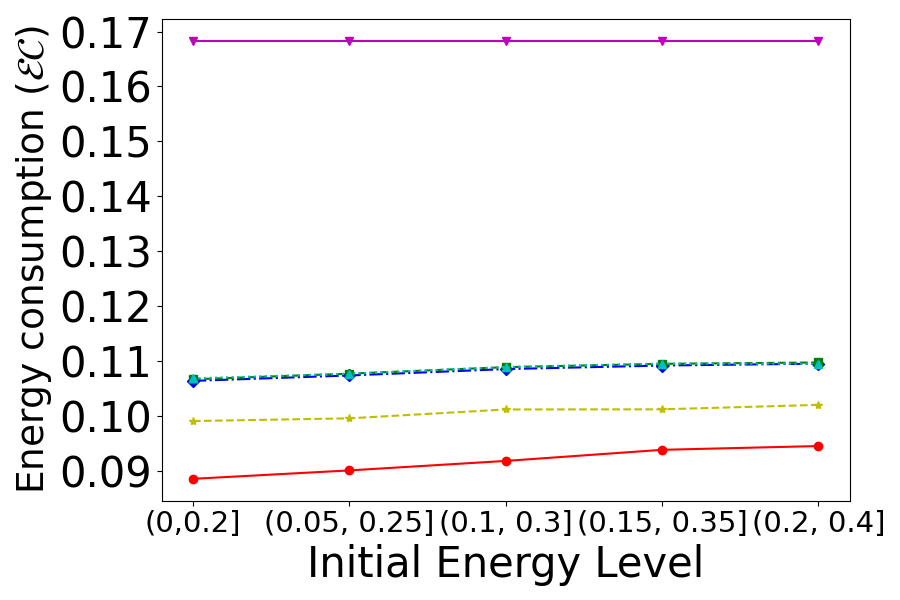}}
  \subfigure[Social Welfare]{
    \includegraphics[width=0.23\textwidth, height=0.2\textwidth]{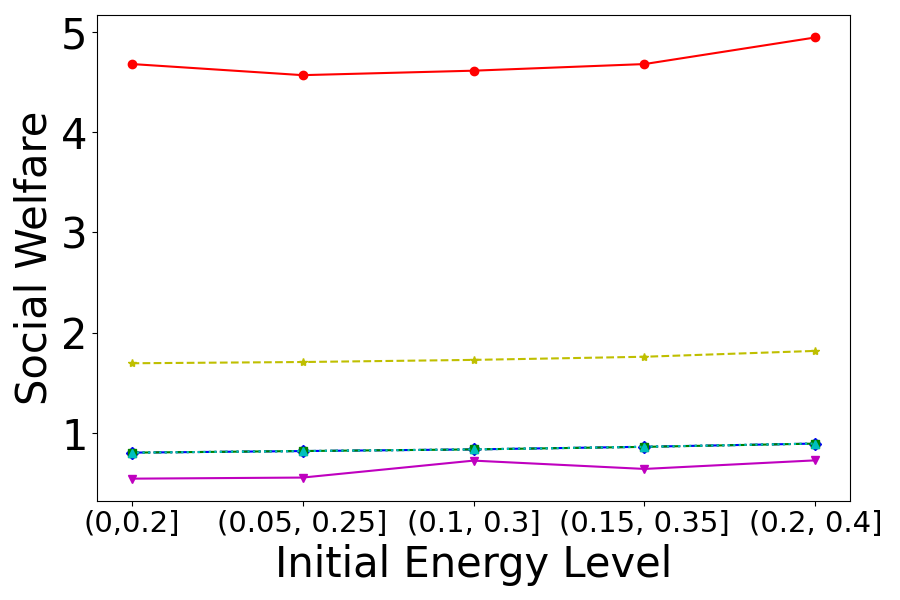}}
    \subfigure[MTBF]{
    \includegraphics[width=0.23\textwidth, height=0.2\textwidth]{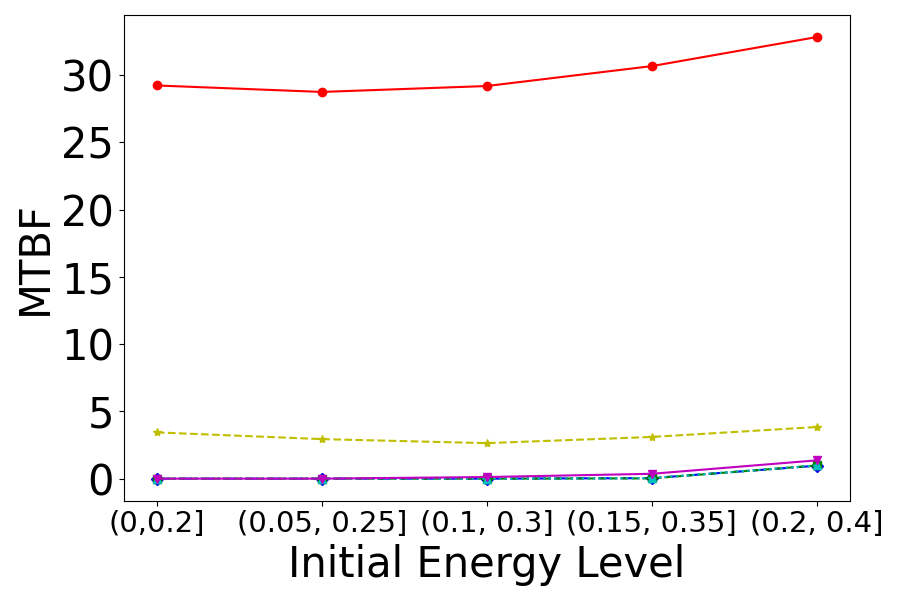}}
    \caption{Effect of Varying Initial Energy Levels ($E_{init}$)}
\label{fig:effect-en}
\vspace{-2mm}
\end{figure*}

Figure~\ref{fig:effect-node} explores how varying initial energy levels in client sensor nodes influence performance metrics. The results indicate that {\tt SusFL} surpasses the considered SOTA schemes in prediction accuracy as the clients' energy levels rise. This advantage stems from {\tt SusFL}'s client selection strategy, prioritizing clients based on their energy availability. Consequently, increasing initial energy allows more clients to engage in FL operations, enhancing prediction accuracy. In contrast, since the number of clients involved in aggregation remains constant in other schemes, their prediction accuracy remains largely unaffected by changes in energy availability.

Furthermore, the presence of more energy in the system, despite causing a slight uptick in energy consumption for the FL process (Figure~\ref{fig:effect-en}(b)), leads to improved social welfare and MTBF as indicated in Figure~\ref{fig:effect-en}(c) and (d). This pattern is unique to {\tt SusFL} as the fixed participation of clients in the FL process does not adjust to the available system energy, highlighting {\tt SusFL}'s adaptability and efficiency in utilizing energy resources for optimal FL operations.

\section{Conclusion \& Future work} \label{sec:conclusion-future-work}
From our research, we highlight the following {\bf key findings} in the performance of {\tt SusFL}:
\begin{enumerate}
\item {\bf Enhanced energy efficiency and prediction accuracy}: {\tt SusFL} demonstrates superior efficiency in global model training within FL operations, utilizing the least amount of energy to attain the highest prediction accuracy compared to benchmark schemes. This efficiency is attributed to an energy-aware client selection mechanism that adeptly chooses an optimal set of clients, thereby balancing high accuracy with energy conservation in the sensor network.

\item {\bf Hightened System availability and MTBF}: {\tt SusFL} excels in MTBF, enhancing system reliability amid energy variability and environmental dynamics in smart farming. This finding emphasizes the scheme's robustness and the necessity to minimize computational demands on sensor nodes, a lesson underscored by {\tt GreenFL}'s relative underperformance.

\item {\bf Improved resilience against attacks}: In scenarios involving cyber and adversarial attacks, {\tt SusFL} stands out for its ability to maintain high prediction accuracy with minimal energy use, thereby ensuring ongoing system availability and sustainability. This resilience highlights {\tt SusFL}'s effectiveness in safeguarding FL operations against potential security threats.
\end{enumerate}
These insights underscore {\tt SusFL}'s comprehensive approach to optimizing FL for energy-constrained environments, offering a scalable, secure, and efficient solution for smart agricultural practices.

For {\bf future work}, we will enhance our study in several key areas:
\begin{itemize}
\item {\bf Fairness and privacy}: Incorporate fairness and privacy preservation into client selection to ensure secure and equitable participation in the FL process.

\item {\bf Scalability enhancement}: Enhance scalability by increasing the number of clients in the FL operations, expanding the system's capacity for larger and more diverse datasets.

\item {\bf Performance optimization}: Optimize performance by refining the client selection mechanism to improve prediction accuracy and energy efficiency, especially for large-scale operations.
\end{itemize}

\newpage
\bibliographystyle{ACM-Reference-Format}
\bibliography{ref}
\end{document}